\documentclass[10pt, conference, compsocconf]{IEEEtran}
\IEEEoverridecommandlockouts
\usepackage{amsmath, amssymb, amsthm, amsfonts}
\usepackage{hyperref}
\hypersetup{colorlinks=false}
\usepackage{color}
\DeclareMathOperator*{\argmin}{arg\,min}
\DeclareMathOperator*{\vecop}{vec}

\usepackage{graphicx}

\begin{document}

\title{HRF estimation improves sensitivity of fMRI encoding and decoding models}

\author{

\IEEEauthorblockN{
Fabian Pedregosa \IEEEauthorrefmark{1}\IEEEauthorrefmark{2}\IEEEauthorrefmark{3}\IEEEauthorrefmark{4},
Michael Eickenberg \IEEEauthorrefmark{1}\IEEEauthorrefmark{2} \IEEEauthorrefmark{3} \thanks{* both authors contributed equally},
Bertrand Thirion\IEEEauthorrefmark{2}\IEEEauthorrefmark{3}
and Alexandre Gramfort\IEEEauthorrefmark{5}\IEEEauthorrefmark{3}}

\IEEEauthorblockA{ \IEEEauthorrefmark{2}Parietal Team, INRIA
  Saclay-\^{I}le-de-France, Saclay, France}

 \IEEEauthorblockA{\IEEEauthorrefmark{3} CEA, DSV,
   I\textsuperscript{2}BM, Neurospin b\^{a}t 145, 91191
   Gif-Sur-Yvette, France}

\IEEEauthorblockA{\IEEEauthorrefmark{4} SIERRA Team, INRIA Paris -
  Rocquencourt, Paris, France} 

\IEEEauthorblockA{\IEEEauthorrefmark{5} Institut Mines-T\'{e}l\'{e}com,
  T\'{e}l\'{e}com ParisTech, CNRS LTCI, Paris, France}

}

\maketitle              

\begin{abstract}
  Extracting activation patterns from functional Magnetic Resonance
  Images (fMRI) datasets remains challenging in rapid-event designs
  due to the inherent delay of blood oxygen level-dependent (BOLD)
  signal. The general linear model (GLM) allows to estimate the
  activation from a design matrix and a fixed hemodynamic response
  function (HRF). However, the HRF is known to vary substantially
  between subjects and brain regions. In this paper, we propose a
  model for jointly estimating the hemodynamic response function (HRF)
  and the activation patterns via a low-rank representation of task
  effects.  This model is based on the linearity assumption behind the
  GLM and can be computed using standard gradient-based solvers. We
  use the activation patterns computed by our model as input data for
  encoding and decoding studies and report performance improvement in
  both settings.
\end{abstract}


\section{Introduction}

The use of decoding models \cite{haynes2006decoding} to predict the
cognitive state of a subject during task performance has become a
popular analysis approach for fMRI studies. The converse approach is the
voxel-based encoding model, which describes the information about the
stimulus or task that is represented in the activity of a single
voxel~\cite{naselaris}.

The input to both types of analysis consists of activation patterns
corresponding to different tasks or stimulus types. These activation
patterns are straightforward to calculate for blocked trials or
slow-event designs, but for rapid-event designs the evoked BOLD
response for adjacent trials will overlap in time, complicating the
identification task.

The general linear model (GLM) was proposed
\cite{friston1994statistical} to overcome this difficulty. It
estimates the activation patterns evoked by separate events and allows
for rapid-event settings. The GLM relies on a known form of the
hemodynamic response function (HRF) to estimate the activation
pattern. However, it is known \cite{handwerker2004variation} that the
shape of this response function can vary substantially across subjects
and brain regions.

In this study we propose to learn the specific form of the HRF in each brain voxel to
improve the computation of the activation vectors in the GLM. Joint
estimation of HRF and activation patterns has already been proposed in
the literature, both within the frequentist \cite{donnet2006fmri} and
Bayesian framework \cite{makni2005joint}. We propose a model based on the linearity assumption behind GLM and low-rank
factorization. We are interested in assessing the impact on
higher-level analysis such as encoding and decoding studies. In
particular, we examine whether encoding and decoding models give
significantly different results when the activation patterns used as
input data are computed using this joint estimation method.

\medskip \textbf{Notation:} $\|\cdot\|$ denotes the euclidean norm for
vectors. $I$ denotes the identity matrix and $e_i$ denotes its $i$th
column vector. $\otimes$ denotes the Kronecker product and \(\vecop(A)\)
denotes the concatenation of the columns of a matrix \(A\) into a
single column vector.

\section{HRF estimation via low-rank approximation}

We denote the observed fMRI time series for a single voxel by $y =
(y_1, \dots , y_n)$ where $y_i$ is the measurement at time $i$TR, with
TR being the time of repetition and $n$ is the number of scans within
the session. As reference HRF we use the one described by Glover
\cite{glover1999deconvolution}, and denote it by $h_c$. In this case,
the GLM model specifies the observed BOLD signal as:
\begin{equation}
\label{eq:glm}
y = X_{h_c} \beta + P w + \varepsilon \enspace,
\end{equation}
where $X_{h_c} \in \mathbb{R}^{n \times p}$ is the design matrix,
i.e. the matrix whose columns are a discrete convolution of $h_c$ with
the binary stimulus vector $V_i$ for condition $i$, $P \in
\mathbb{R}^{n \times q}$ is the matrix of confounds (drifts, motions
etc.)  and $\varepsilon \in \mathbb{R}^n$ is the vector of residuals,
which is modeled as an auto-regressive process of order 1 (AR1(1)) to
take into account the temporal correlations in the noise. We now
define the binary matrix ${\tilde{X} \in \mathbb{R}^{n \times p r}}$
as
\begin{equation}
\label{eq:design}
\tilde{X} = \sum_{k=0}^r L_k (V \otimes e_k^T)
\end{equation}

\begin{figure}[!t]
\centering
\includegraphics[scale=.25]{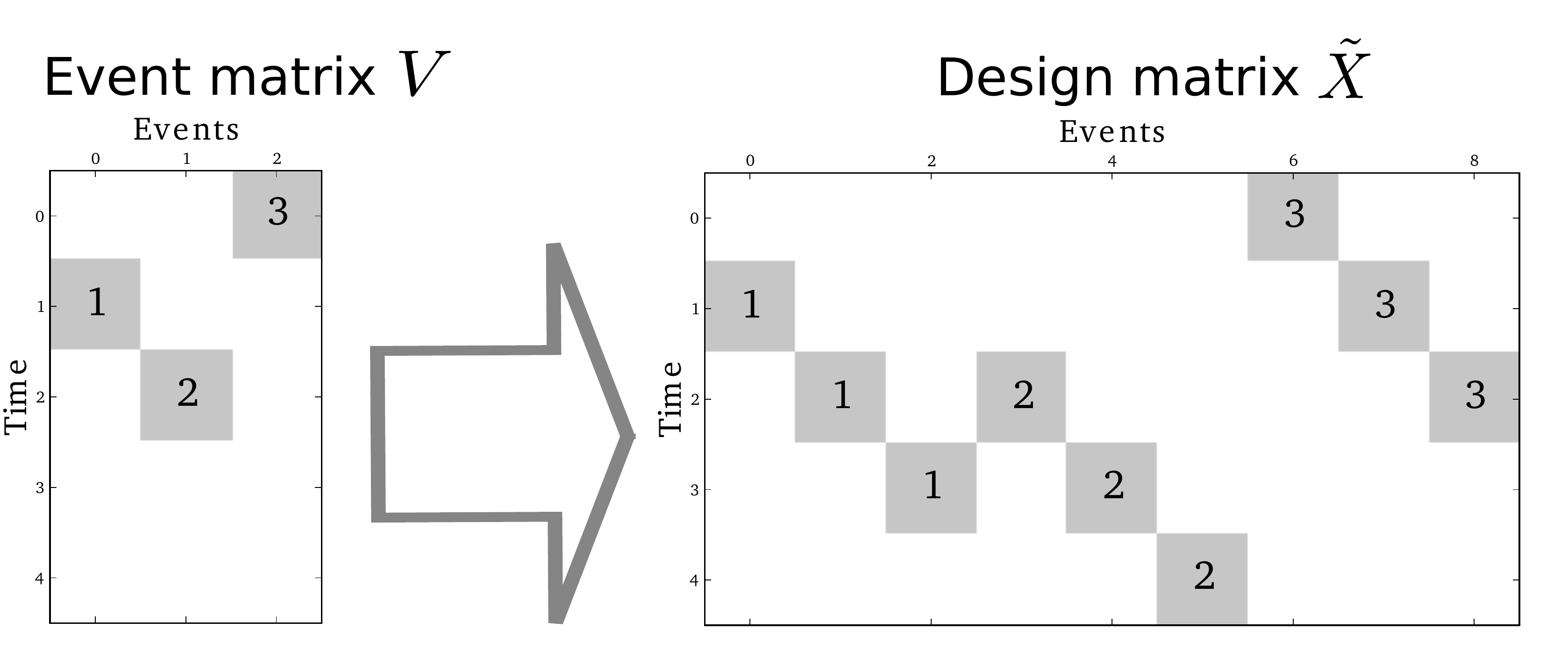}
\caption{Design matrix for Finite Impulse Response models. Different numbers
  correspond to different events. Here the HRF is assumed to span
  over a period of 3 TRs.}
\label{fig:design}
\end{figure}
where $r$ is the duration of the HRF in multiples of TR, $V$ the
binary stimulus matrix and $L_k$ is the lower shift matrix of order
$k$, i.e. the matrix that shifts downwards $k$ places all
elements. Figure~\ref{fig:design} illustrates the relationship between
the event matrix and the design matrix. Matrix $\tilde{X}$ has the
property that ${\tilde{X} \vecop{(h \beta^T)}} = X_h \beta$, for any
vector \(h\), allowing us to write equation \eqref{eq:glm} with an
explicit linear dependency on $h$, $y = \tilde{X} \vecop (h_c \beta^T)
+ P w + \varepsilon$.

The model parameters are then estimated such that they minimize the
decorrelated residuals. Equivalently, we solve the following
optimization problem for $\beta$, $h$ and $w$:
\begin{equation}
\label{eq:obj}
\argmin_{\beta, h, w} \frac{1}{2}\|Z^T(y - \tilde{X} \vecop(h \beta^T) - P w)\| ^2 \enspace ,
\end{equation}
where \(Z^T\) is the whitening matrix associated with the AR(1)
covariance matrix     
estimated by maximum likelihood.  This represents a linear regression
model where the coefficients are the vectorization of a
\mbox{rank-one} matrix. It can also be seen as a Finite Impulse
Response (FIR) basis GLM with a constraint such that the HRF $h$ is repeated
for each condition. We will refer to this as the {\it rank-one
  regression} model. Note that while it is possible to add a
regularization term to reflect prior knowledge of the shape of the
HRF, no specific form of $h$ is privileged within this model.


\subsection{Choice of a particular basis}

In certain cases it may be desirable to reduce the number of free
parameters of the HRF function within the rank-one model. An effective
technique consists in constraining $h$ to be obtained as a linear
combination of $t$ basis functions with $t \leq r$, in an approriately
chosen basis. Common bases used to express the HRF include the
Fourier basis, polynomial basis and the canonical HRF together with
its derivatives up to a certain order.

Given the subspace generated by the columns of a matrix $Q \in \mathbb{R}^{r \times t}$,
we impose $h$ to lie in that subspace. In that case we can write $h = Q \alpha$
for some $\alpha \in \mathbb{R}^t$. By the properties of the Kronecker 
product we have the equivalence ${\tilde{X} \vecop(Q \alpha \beta^T)} =
{\tilde{X} (I \otimes Q)   \vecop(\alpha \beta^T)}$. We can now define our
rank-one model over parameters $\alpha$ instead of $h$ by defining the design
matrix $\tilde{X}_Q = \sum_{k=0}^t L_k (V \otimes q_k^T)$ where $q_i$ denotes
the $i$th column of $Q$.

\subsection{Asynchronous design}

When the events are not a multiple of the repetition time,
constructing the design matrix $\tilde{X}$ cannot be accomplished
directly by equation \eqref{eq:design}.

Let $V$ represent the binary stimulus where the events are truncated
to the closest TR. Furthermore, let $V = \sum_{i=0}^m E_i$, where
$E_i$ is the matrix of individual events.  In the case where a
continuous basis for the HRF is chosen this can be evaluated at all
time. That is, the design matrix can be written as

$$
\tilde{X} = \sum_{i=0}^m \sum_{k=0}^t L_k (E_i \otimes q_{k, i})
$$

where $q_{k, i}$ represents the $k$th vector in the basis evaluated at
the TR timepoints plus an offset equal to the offset of event $i$ with
respect to the TR.

\subsection{Algorithm}

Although the problem \ref{eq:obj} is not convex, the cost function is
differentiable and \mbox{gradient-based} methods can be used to solve the
optimization problem. We used the limited-memory BFGS algorithm~\cite{lbfgs}
to simultaneously optimize over parameters $\beta, h, w$. For simplicity, we
have taken $Z$ to be the identity. For the general result, simply multiply $X,
y, P$ by $Z^T$. Popular implementations of the algorithm only need as
parameters the objective function \eqref{eq:obj} and the gradient, given by:
\begin{align*}
\nabla_\beta =& (I \otimes h^T) \tilde{X}^T (\tilde{X} \vecop(h \beta^T) + P w - y) \\
\nabla_h =& (\beta^T \otimes I) \tilde{X}^T (\tilde{X} \vecop(h \beta^T) + P w - y) \\
\nabla_w =&  {P}^T (\tilde{X} \vecop(h \beta^T) + P w - y)
\end{align*}

The full gradient now can be computed by stacking $\nabla_\beta,
\nabla_h$ and $\nabla_w$ into a single vector. Since in this algorithm
only a matrix-vector product is used, the use of Kronecker product identities
avoids the explicit creation of most Kronecker product matrices. We
have found this implementation to take around 3 hours to perform a
full brain analysis (124.000 voxels, 46 conditions) on commodity
hardware with almost no increase in memory consumption once the data
was loaded in memory. An implementation by the authors is publicly
available \footnote{\href{https://pypi.python.org/pypi/hrf\_estimation}{https://pypi.python.org/pypi/hrf\_estimation}}

\begin{figure}[!t]
\centering
\includegraphics[scale=.2]{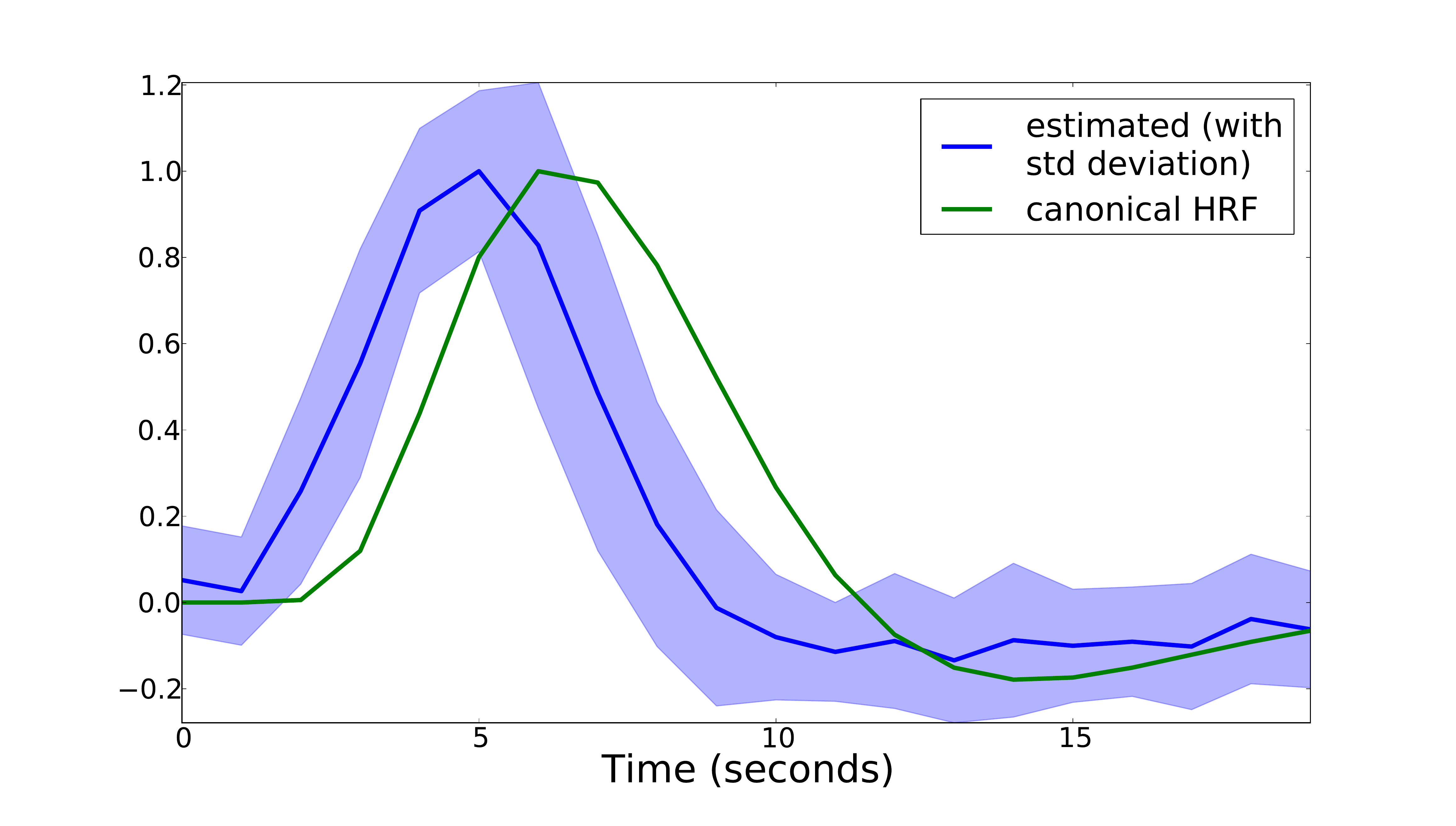}
\includegraphics[scale=.2]{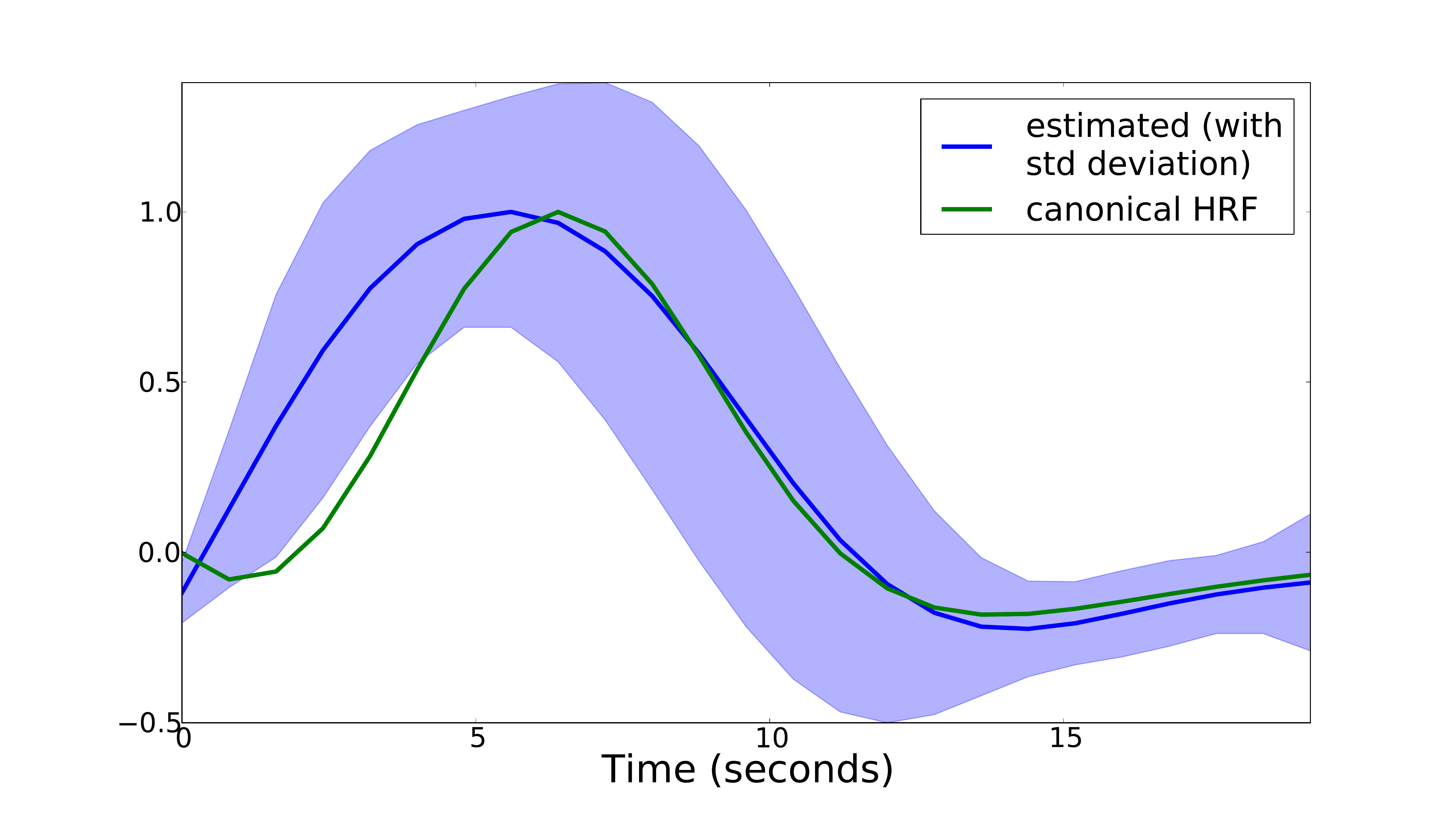}
\caption{ Estimated HRF for the two different datasets, using the top
  100 performing voxels. In blue we show the mean HRF across these
  voxels, together with its standard deviation in more transparent
  color. For the first image, the HRF was expressed using a FIR basis
  with 20 degrees of freedom. For the second images, a basis of 6
  elements consisting on the canonical HRF and its five succesive
  derivatives was chosen.}
\label{fig:hrf_comparison}
\end{figure}

\section{Validation}

We use two fMRI datasets (one for encoding, one for decoding), and for
each we consider two validation criteria. The first criterion is
common to both datasets and assesses whether the learned HRF fits
better unseen data than the canonical function. We fit a
\mbox{rank-one} regression model on all but one session. This gives us
an estimate of the HRF for all voxels. We then use this HRF to compute
the log-likelihood of a GLM model on unseen data. In similar fashion,
we compute the log-likelihood using the canonical HRF. A paired test
allows us to conclude whether both likelihoods are significantly
different and thus if one model has better goodness of fit.

This criterion serves to validate our model using a likelihood
function defined on the raw fMRI timeseries.  However, we are mainly
interested in the GLM as a pre processing step for higher level
analysis such as decoding or encoding models. That is why, in order to
fully assess the relevance of this contribution, we propose to
quantify the performance of our rank-one model using the performance
metrics commonly used with decoding and encoding models.


\subsection{Data description}

{\it{Natural images}} For the encoding model, we use the publicly
available dataset from~\cite{kay2008identifying}, where the task
consists in predicting the activation maps from a set of natural
images. Subjects viewed 1750 training images, each presented twice,
and 120 validation images, each presented 10 times, while fixating a
central cross. Images were flashed 3 times per second (200ms
on-off-on-off-on) for one second every 4 seconds, leading to a rapid
event-related design. The data was acquired in 5 scanner sessions,
each comprising 5 blocks of 70 training images, each presented twice
within the block and 2 blocks of validation images showing 12 images
each 10 times.

Evaluation of the performance of our GLM was done with a simple
encoding task: Using a spatially smoothed Gabor pyramid transform
modulus with 2 orientations and 4 scales, we used Ridge regression to
learn a predictor of voxel activity on 80\% (4 of 5 sessions,
i.e. 1400 images) of the training set. Linear predictions on the left
out fold were compared to the true activations in \(l_2\) norm and
normalized by the variance of the voxel activity, before being
subtracted from 1 (predictive \(r_2\) scoring).

{\it{Word decoding}} For the decoding task, we use the dataset
described in \cite{gramfort-etal:2012b}, where the task consist in
predicting the visual percept formed by four letter words. Each word
was presented on the screen for 3\,s at a flickering frequency of
15\,Hz. A 5\,s rest interval was inserted between each word
presentation. The subject was asked to fixate a colored cross at the
center of the screen. Each session comprised 46 words including 6
verbs. To ensure that subjects were reading, they were asked to report
with a button press when a verb was presented on the screen.
Repetitions corresponding to verbs were then removed from the
analysis. Six acquisition blocks were recorded, leading to 240
different words used in the analysis. We evaluate the performance of
our GLM in this decoding task by calculating the percentage of overall
correctly predicted bars forming the image.

\subsection{Results}

We first report results on the natural images dataset. As described
above, we validate by fitting a rank-one regression model on all but
one session. On the left-out session, we found the log-likelihood of
the GLM model obtained with the data-driven HRF to be consistently
larger than the log-likelihood obtained using a canonical HRF. A
paired difference test was used to conclude that the mean likelihood
of the rank-one model is significantly larger with \mbox{p-value $<
  10^{-3}$}.

The estimated HRF across the 100 most responding voxels is presented in
Fig.~\ref{fig:hrf_comparison}. We show the average value for the
learned HRFs across voxels together with its standard deviation. As
expected, the data-driven HRFs resemble the canonical HRF, however,
the peak is located on average one second before the peak of the
canonical function.

\begin{figure}[!t]
\centering
\includegraphics[scale=.43]{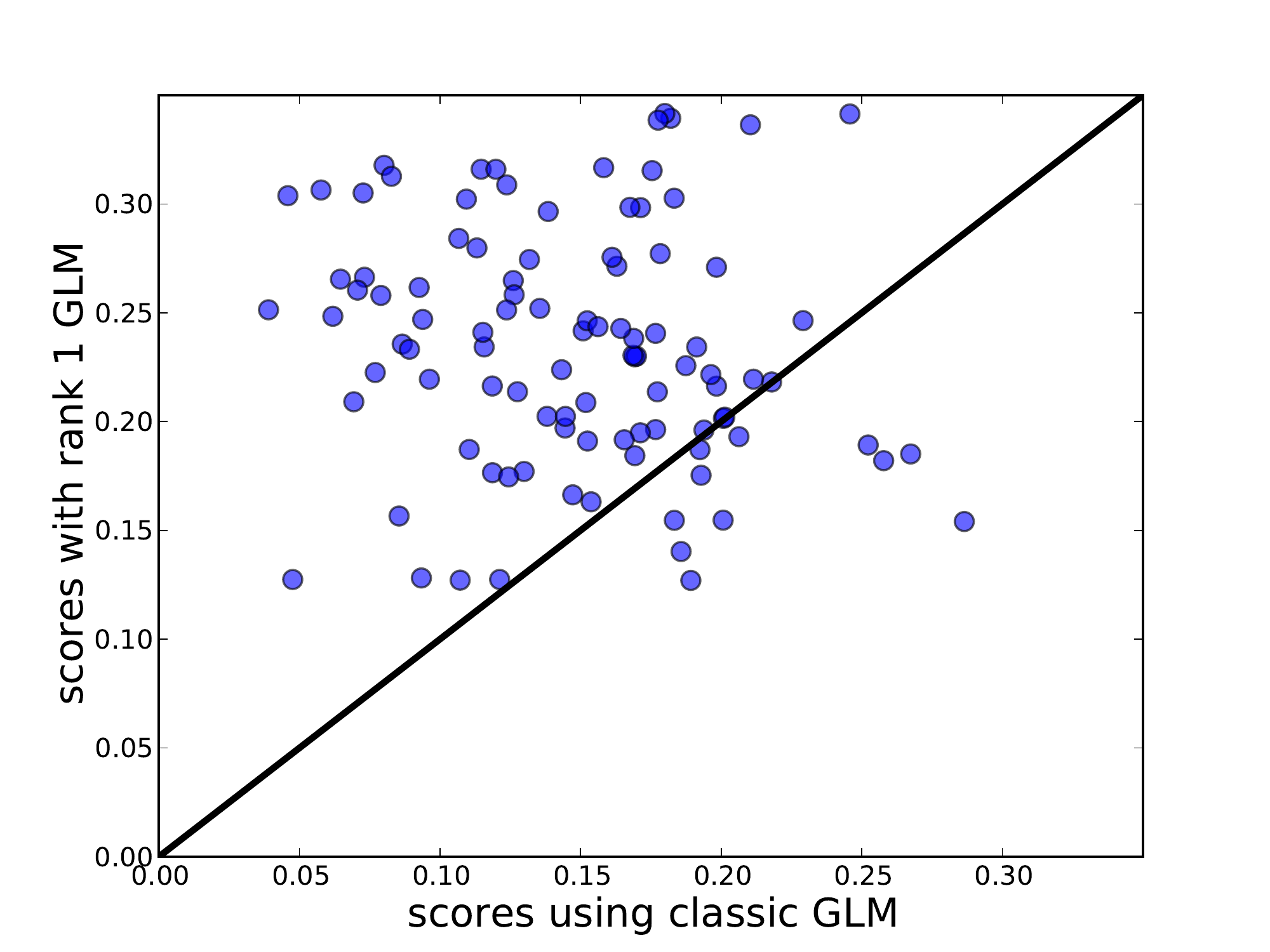}
\caption{Performance scores of individual voxels using the rank-one
  model against the performance obtained by a standard GLM. Elements
  over the diagonal represent voxels in which our model obtained
  higher scores.}
\label{fig:michael_score}
\end{figure}

The predictive \(r_2\) scores of 100 voxels are shown in a scatter
plot in Fig.~\ref{fig:michael_score}. We chose the 100 best predicted
voxels using a classic GLM. These scores are on the x-axis. The y-axis
shows the scores using our method. Significantly more scores lie above
the diagonal (\(p < 10^{-4}\)), Wilcoxon signed rank test), suggesting
that that learning the HRF is beneficial to this encoding scheme.

The second dataset that we analyze is the \textit{word-decoding}
dataset. This dataset is an asynchronous setting with TR = 2.4\,s and
trials of 8\,s. As described above we chose to oversample by a factor
or 3 the original TR to a TR of 0.8s. Because of the low time
resolution and the shorter time course, we chose to constrain the HRF
to a small subspace in order to avoid overfitting associated with
complex models. The set of basis functions for this subspace is given
by the canonical HRF and its five successive derivatives. We have
experimented with several different bases including Chebyshev
polynomials, Fourier basis and discrete cosine basis. All these basis
capture the general trend of the HRF function and give similar
results, but each set induces some bias towards specific shape
functions, while this set of generators favors more biologically
plausible models. As with the previous dataset, we observed
log-likelihood values that are constantly larger than for the GLM. A
paired difference test was used to conclude that the mean
log-likelihood is significantly higher with \mbox{p-value $< 3 \times
  10^{-3}$}.


As can be seen in Fig.~\ref{fig:hrf_comparison} second image, the
estimated HRF resembles the canonical HRF, which is not surprising
given the subspace in which it is constrained. As with the previous
dataset, the peak of the HRF is slightly advanced with respect to the
canonical function.

We used the activation patterns estimated by the rank-one regression
model as input to the decoding study. Within the four subjects
analyzed, we observed a systematic increase in the mean score, ranging
for +1\% to +3\% (out of a score of ~65\%). We observed higher
increase for the best performing subjects. A Wilcoxon signed-rank test
on the scores across subjects gave us a \mbox{p-value $< 0.07$} for
the significance of these differences.

\section{Conclusion}

We have presented a model that jointly estimates the hemodynamic
response function (HRF) and the activation patterns from the BOLD
signal. This model, named \mbox{\it rank-one} regression,
can be optimized using standard smooth optimization methods such as
L-BFGS. We investigated whether this model yields better prediction
for encoding and decoding models.

In a first step, we assessed the quality of the HRF estimation by
comparing the likelihood of the GLM on unseen data using both the
estimated HRF and the canonical HRF. To assess the impact on
encoding and encoding models, we have selected two fMRI datasets
and used the GLM coefficients obtained by our rank-one model as
input data. 

We found out that using the activation patterns estimated by the
rank-one model significatively improved encoding and decoding
studies. In the encoding study we found a generalized improvement
across voxels. In the decoding study observed improved scores for all
subjects across the study.

\section{Acknowledgment}

This work was supported by grants IRMGroup ANR-10-BLAN-0126-02 and
BrainPedia ANR-10-JCJC 1408-01. The authors would like to thank
Guillaume Obozinski, Francis Bach and Philippe Ciuciu for fruitful
discussions.

%
%
\bibliographystyle{IEEEtran}
\bibliography{biblio}{}
\end{document}